\documentclass[10pt,twocolumn,letterpaper]{article}

\usepackage[pagenumbers]{cvpr}

\usepackage{graphicx}
\usepackage{amsmath}
\usepackage{amssymb}
\usepackage{booktabs}
\usepackage{textcomp}

\usepackage[pagebackref,breaklinks,colorlinks]{hyperref}
\usepackage[capitalize]{cleveref}

\expandafter\def\expandafter\normalsize\expandafter{%
	\normalsize
	\setlength\abovedisplayskip{8pt}
	\setlength\belowdisplayskip{8pt}
	\setlength\abovedisplayshortskip{8pt}
	\setlength\belowdisplayshortskip{8pt}
}

\crefname{section}{Sec.}{Secs.}
\Crefname{section}{Section}{Sections}
\Crefname{table}{Table}{Tables}
\crefname{table}{Tab.}{Tabs.}

\def\confName{CVPR}
\def\confYear{2022}

\begin{document}

\title{OccAM's Laser: Occlusion-based Attribution Maps \\ for 3D Object Detectors on LiDAR Data}

\author{
	{David Schinagl\textsuperscript{1,2} \quad 
		Georg Krispel\textsuperscript{1} \quad 
		Horst Possegger\textsuperscript{1} \quad 
		Peter M. Roth\textsuperscript{3,4} \quad 
		Horst Bischof\textsuperscript{1,2}} \\ 
	{\tt\small\{david.schinagl,georg.krispel,possegger,bischof\}@icg.tugraz.at}, \quad
	\tt\small{peter.roth@tum.de}\\
	\textsuperscript{1} Graz University of Technology \quad 
	\textsuperscript{2} Christian Doppler Laboratory for Embedded Machine Learning  \\
	\textsuperscript{3}	Technical University of Munich \quad
	\textsuperscript{4} University of Veterinary Medicine, Vienna
}
\maketitle

\begin{abstract}
	While 3D object detection in LiDAR point clouds is well-established in academia and industry, the explainability of these models is a largely unexplored field.
	In this paper, we propose a method to generate attribution maps for the detected objects in order to better understand the behavior of such models.
	These maps indicate the importance of each 3D point in predicting the specific objects.
	Our method works with black-box models: We do not require any prior knowledge of the architecture nor access to the model's internals, like parameters, activations or gradients.
	Our efficient perturbation-based approach empirically estimates the importance of each point by testing the model with randomly generated subsets of the input point cloud.
	Our sub-sampling strategy takes into account the special characteristics of LiDAR data, such as the depth-dependent point density.
	We show a detailed evaluation of the attribution maps and demonstrate that they are interpretable and highly informative.
	Furthermore, we compare the attribution maps of recent 3D object detection architectures to provide insights into their decision-making processes. 
\end{abstract}
\section{Introduction}
\label{sec:intro}

\begin{figure}[t]
	\centering
	\includegraphics[width=0.85\linewidth]{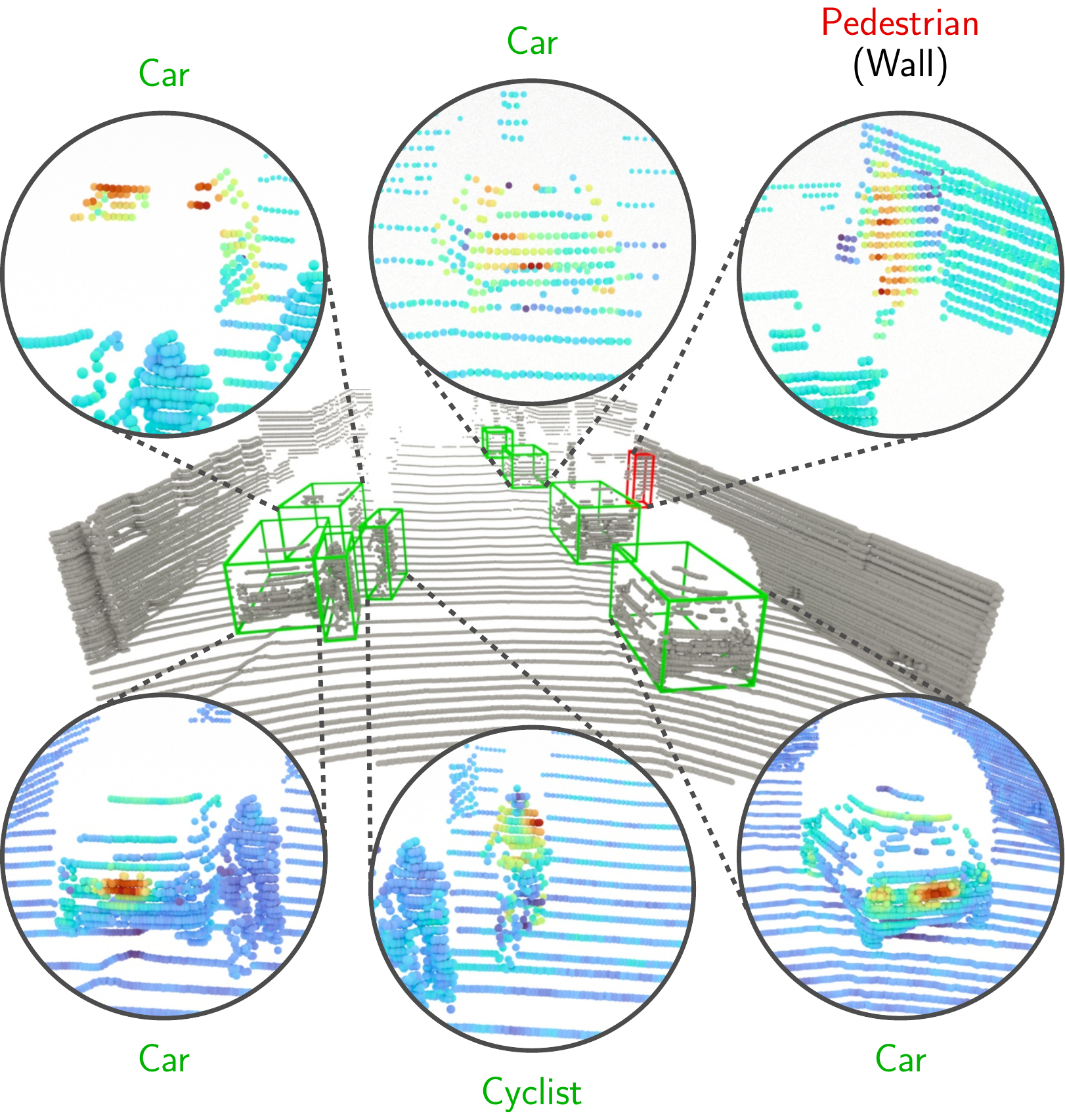}
	\caption{Our attribution maps show the importance of individual LiDAR points for the black-box detection model's decisions.}
	\label{fig:intro}
\end{figure}

Driven by the race to autonomous driving, the interest in 3D object detection is increasing in both, industry and the research community.
Especially the detection of objects in point clouds acquired with LiDAR~(Light Detection and Ranging) sensors is of great interest.
On the one hand, this can be seen by the number and volume of published datasets that include LiDAR data,  \eg~\cite{geiger2012CVPR,sun2020CVPR,chang2019CVPR,caesar2020CVPR,huang2020PAMI} and, on the other hand, by the number of recent publications that deal with this topic, \eg~\cite{yang2018CVPR, yang2020CVPR, wang2020ECCV, he2020CVPR, ye2020CVPR, shi2019CVPR, zhou2018CVPR, yan2018Sensors, lang2019CVPR, shi2020CVPR:2, yang2019ICCV, shi2021PAMI, yin2021CVPR, mao2021ICCV:2}.
In contrast, the explainability of these detectors based on deep neural networks is a rather unexplored area. 
For a safety-relevant area such as autonomous driving, however, it is of particular importance to make the decisions of these inherently opaque models more transparent.

An established technique for the analysis of models based on deep neural networks operating on image data is the generation of attribution or saliency maps, \eg~\cite{simonyan2013arxiv, zeiler2014ECCV, sundarajan2017ICML, zeiler2014ECCV, springenberg2015ICLRW, selvaraju2017ICCV}.
The goal is to visualize the importance of input pixels on the model's decision-making process. 
These have been proven useful for interpreting detections or analyzing the learned features. 
However, for models operating on point clouds there are only a few approaches that address the generation of point-level attribution maps, \eg~\cite{gupta2020IJCNN, zheng2019ICCV}. 
These focus exclusively on the analysis of point cloud \emph{classification} models. 
The analyzed models operate on artificially generated point clouds derived from CAD models, such as in the ModelNet dataset~\cite{wu2015CVPR}. 
Moreover, these attribution map approaches assume prior knowledge of the architecture and require access to the gradients of the model. 
Thus, they are only applicable for the analysis of white-box classification models. 
These properties prevent using them for the analysis of 3D object detectors on LiDAR point clouds. 
First, for 3D object detection, not only the \emph{classification} of the object is important, but also its \emph{localization}. 
These tasks usually go hand in hand and cannot be analyzed separately. 
Second, the detectors typically consist of complex architectures in which different parts contribute to the final detection, \eg~\cite{shi2021PAMI, shi2020CVPR:2}, making gradient-based methods difficult or infeasible. 
In addition, a  non-differentiable pre-processing step used in some models, \eg~\cite{lang2019CVPR, yang2018CVPR}, prevents the backpropagation to the input points.

To the best of our knowledge, we present the first method to generate point-level attribution maps for 3D object detectors on LiDAR point clouds. 
For each detected object a separate map is created, as shown in Figure~\ref{fig:intro}, which reflects the influence of each 3D point on the classification and localization of the object. Our method is applicable to black-box models and thus, in principle, to any detector, since neither prior knowledge of the model's architecture nor access to its internals is required. 

Inspired by perturbation approaches to generate saliency maps for image-based black-box models~\cite{zeiler2014ECCV, petsiuk2018BMVC, petsiuk2021CVPR}, we leverage the principle of \emph{analysis by occlusion}. 
We propose \emph{OccAM:} \textbf{Occ}lusion-based \textbf{A}ttribution \textbf{M}aps for 3D object detectors on LiDAR data. 
We estimate the importance of points by testing the model with randomly generated subsets of the input point cloud. 
The underlying assumption is that an object will be detected less accurately or not at all if areas of the input that are important for the detection have been removed or perturbed.
However, the special characteristics of point clouds, like the unstructured nature and the depth-dependent point density in LiDAR data, poses new challenges compared to the analysis of image-based models. 
Thus, we propose a voxel-based sub-sampling strategy in which the sampling probability is adapted to the point density to challenge the detector appropriately. 
To evaluate the influence on the detections as precisely as possible, we further use a similarity metric that is optimized for the properties of 3D bounding boxes.

In a detailed analysis for PointPillars~\cite{lang2019CVPR}, we demonstrate that our attribution maps are interpretable and informative, \eg~allowing us to analyze the potential source of false detections. 
We show that by averaging multiple maps of a class, regions of particular importance can be derived. 
By comparing the average attribution maps of different state-of-the-art 3D object detectors, \ie~PointPillars, SECOND~\cite{yan2018Sensors} and PV-RCNN~\cite{shi2020CVPR:2}, we further analyze whether differences can be detected.

\section{Related Work}
\label{sec:rel_work}

\noindent\textbf{LiDAR-based 3D Object Detection:}
These methods can be roughly categorized into point-based, grid-based, and hybrid methods depending on how they handle the unstructured format of point clouds. 
\textbf{Point-based methods} directly process the unordered points~\cite{yang2020CVPR, shi2020CVPR, Li2021CVPR, qi2017CVPR, qi2017NIPS, shi2019CVPR}. 
PointNet~\cite{qi2017CVPR} and its successor PointNet++~\cite{qi2017NIPS} showed in their pioneering work a way to directly extract features using shared MLPs along with a global pooling function. 
In PointRCNN~\cite{shi2019CVPR}, PointNet++ features are used to segment foreground points and to generate proposals. 
\textbf{Grid-based methods}, on the other hand, discretize the 3D space into regular grids that can then be processed more easily~\cite{yang2018CVPR, wang2020ECCV, zhou2018CVPR, yan2018Sensors, lang2019CVPR}. 
In the end-to-end trainable VoxelNet~\cite{zhou2018CVPR}, the 3D space is divided into equally sized voxels where the features within each voxel are learned using PointNet. 
To increase the efficiency of VoxelNet, an improved 3D sparse convolution is used in SECOND~\cite{yan2018Sensors}. 
Instead of using voxels, PointPillars~\cite{lang2019CVPR} introduced a column (pillar) representation to further reduce the complexity. 
\textbf{Hybrid methods} such as PV-RCNN~\cite{shi2020CVPR:2, shi2021arxiv} or STD~\cite{yang2019ICCV} simultaneously process point and voxel information to obtain multi-scale features and fine-grained localization~\cite{he2020CVPR, ye2020CVPR, miao2021CVPR, noh2021CVPR}. 
For example, Part-$\text{A}^2$~\cite{shi2021PAMI} uses point-wise features to predict intra-object parts and then aggregates the part information for box refinement to improve the robustness. 
Recently, several approaches based on transformers~\cite{vaswani2017NIPS} have been proposed using point-based and / or grid-based features~\cite{sheng2021ICCV, mao2021ICCV, guan2021arxiv, pan2021CVPR}.

\noindent\textbf{Attribution Maps for Image-based Models: }
To analyze \textbf{white-box} models, where access to the internals of the model is possible, several gradient-based approaches were proposed, which leverage the gradient of the output with respect to the input image~\cite{simonyan2013arxiv}. 
Due to the gradient backpropagation, however, these approaches tend to create noisy maps. 
Hence, some methods average multiple gradients~\cite{sundarajan2017ICML, bach2015Plos, smilkov2017arxiv} or try to improve the backpropagation using gradient calculation rules~\cite{zeiler2014ECCV, springenberg2015ICLRW, ruiz2021ICCV, shrikumar2017ICML, zhang2018IJCV}.
Fong~\etal~\cite{fong2017ICCV, fong2019ICCV} create a minimal perturbation mask that minimizes the probability of the target class by backpropagating the error.
Another type of methods for white-box analysis uses the activations of intermediate layers~\cite{rebuffi2020CVPR, zhou2016CVPR, selvaraju2017ICCV, bakken2020ECCV}. 
For instance, Grad-CAM~\cite{selvaraju2017ICCV} uses the spatial information available in convolutional layers and weights the activation scores at every location with the gradients of the class score \wrt the layer. 
Bakken~\etal~\cite{bakken2020ECCV} use PCA to decompose the features of the final convolutional layer.

On the contrary, to analyze \textbf{black-box} models neither prior knowledge about the architecture nor access to the internals is available. 
These methods generate the attribution map by perturbing the input and analyzing the effects on the model output~\cite{zintgraf2018ICLR, lundberg2017NIPS} and are closely related to the Shapley Value~\cite{shaple1952}, a concept from game theory where the goal is to determine a player's contribution to the final outcome. 
Zeiler and Fergus~\cite{zeiler2014ECCV} use a sliding window to block out input patches while observing the drop in the output score of a classification model. 
The superpixel-based  LIME~\cite{ribeiro2016SIGKD} uses a linear surrogate model that estimates the impact on the output if the superpixels are removed. 
RISE~\cite{petsiuk2018BMVC} averages random binary masks according to the model's output class probability for the masked inputs. 
This is extended in D-RISE~\cite{petsiuk2021CVPR} by a similarity metric allowing its application to detection models as well.

Overall, black-box analysis is computationally expensive, but of course provides the advantage of more general applicability. 
Inspired by the insights of these explanation methods for image data, we address the creation of attribution maps for black-box 3D object detectors to better understand their decision-making processes.

\noindent\textbf{Attribution Maps for Point Cloud-based Models:} In contrast to attribution maps for image-based models, there are only few approaches that investigate point cloud-based models. 
The existing methods focus on the analysis of white-box classification models, mainly based on PointNet architectures that process synthetic point clouds, like in the ModelNet dataset~\cite{wu2015CVPR}. 
Gupta~\etal~\cite{gupta2020IJCNN} adapt gradient-based attribution maps for image classification models to analyze point cloud classification models trained on ModelNet data.
Zheng~\etal~\cite{zheng2019ICCV} propose a gradient-based method based on the idea of approximating the non-differentiable point dropping by moving the points towards the point cloud center. 
The underlying assumption is that points in the center are uninformative for the classification model and thus, point dropping can be approximated by this differentiable operation. 
However, this is only true for point clouds derived from CAD models like in the ModelNet dataset. 
In real LiDAR data, there are no points on back-facing surfaces of an object, but potentially important points within the object. 
For example, a driver sensed through the side window of a car  can be a cue for the detector.

Distantly related are adversarial attacks on point cloud classification models. 
More specifically, methods that try to identify the most important points for the model in order to target them~\cite{liu2019ICIP, wicker2019CVPR, zhang2019arxiv, kim2021ICCV}. 
However, they also only target white-box classification models, trained almost exclusively on ModelNet data, and determine the saliency using gradient-based methods.

In contrast to existing approaches, we are able to generate attribution maps for 3D object detection models that process wide-range LiDAR point clouds. 
Since the goal of 3D object detection is not only the classification of a single object, but the localization and classification of multiple objects, these detectors often consist of complex architectures, \eg~\cite{shi2021PAMI, shi2020CVPR:2}. 
This makes gradient-based analysis difficult or infeasible at all. 
In addition, the models often use a non-differentiable pre-processing step, \eg~\cite{lang2019CVPR, yang2018CVPR}, preventing the backpropagation to the 3D points. 
Therefore, inspired by model-agnostic approaches for image-based models~\cite{zeiler2014ECCV, petsiuk2018BMVC, petsiuk2021CVPR}, we propose a method to generate attribution maps for black-box 3D object detection models.

\section{Occlusion Analysis for LiDAR Data}
\label{sec:method}

\begin{figure*}
	\centering
	\includegraphics[width=0.98\linewidth]{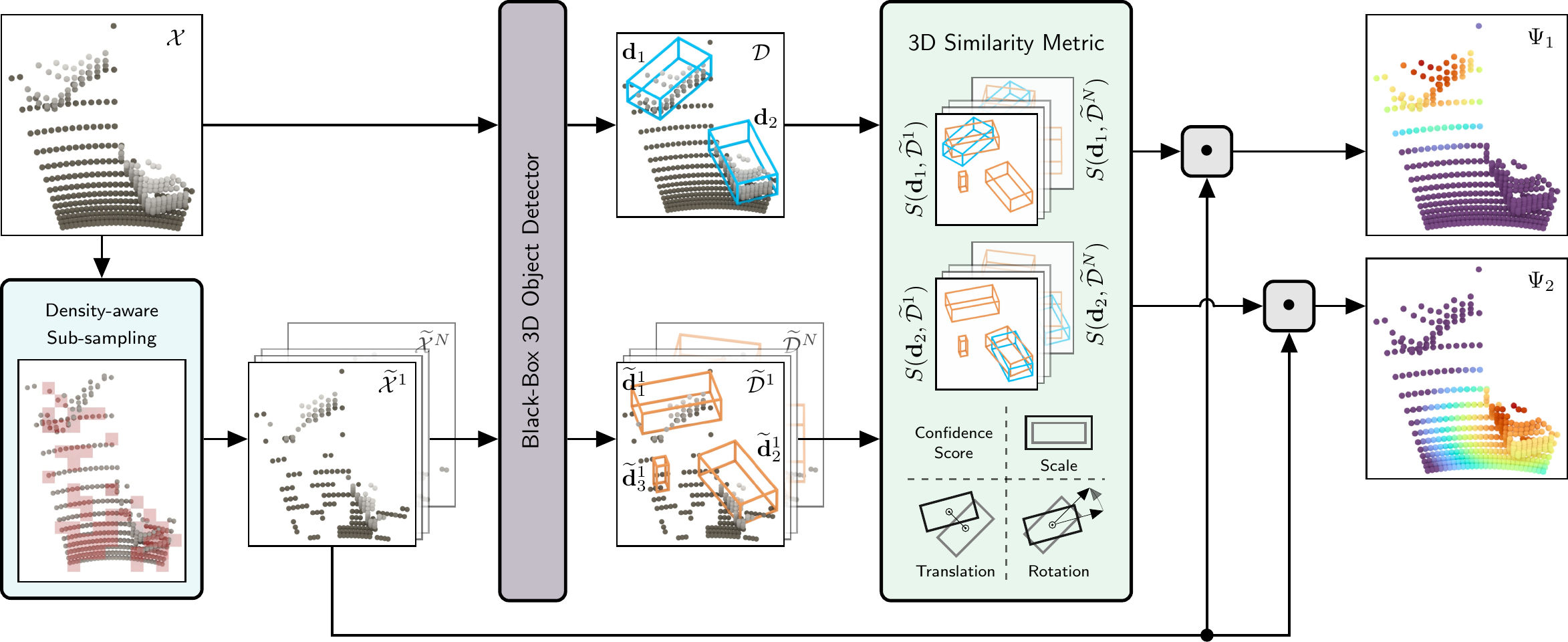}
	\caption{OccAM overview. We consider the specific point cloud characteristics during the sub-sampling of the input point cloud, such that the black-box 3D object detectors are challenged appropriately for all sensing ranges. 
		Using a similarity metric tailored for this task, we can precisely analyze the changes in the detection output to obtain highly expressive attribution maps.}
	\label{fig:overview}
	\vspace{-0.4cm}
\end{figure*}

Our goal is to create an attribution map for each detected object in a point cloud. 
The map should indicate the importance of each 3D point for the detector's prediction regarding the specific object. 
We do not require any knowledge about the architecture of the model, nor access to parameters, features or gradients. 
This allows universal applicability and thus, the comparison of different architectures independent of their design. 
OccAM provides insights into the decisions of a black-box detector by leveraging occlusion analysis. 
The basic idea is to estimate the importance of input points by systematically removing them while observing changes in the output of the model being analyzed, as illustrated in Figure~\ref{fig:overview}. 
Compared to methods that have successfully applied this principle to analyze 2D image models~\cite{zeiler2014ECCV,  petsiuk2018BMVC, petsiuk2021CVPR}, analyzing LiDAR-based 3D object detectors poses the following new challenges, which we address:
\begin{itemize}
	\item The unstructured format of point clouds makes the systematic sub-sampling of the input data more complex. 
	We propose an efficient voxel-based sampling approach that allows masking of adjacent regions while preserving point-level information (Section~\ref{sec:method:voxel-masking}).	
	\item An inherent characteristic of LiDAR data is the non-uniform depth-dependent point density which must be taken into account during sub-sampling. 
	Hence, we propose a density-aware sampling strategy to allow simultaneously analyzing all objects regardless of their distance to the LiDAR sensor (Section~\ref{sec:method:density-aware}).
	\item Intersection over Union (IoU) alone is not sufficient to evaluate subtle detection changes precisely. 
	Therefore, we use a similarity metric that evaluates translation, rotation and scaling individually (Section~\ref{sec:method:metric}).
\end{itemize}

\noindent\textbf{Definitions \& Notations: } Let ${\mathcal{X} = \{\mathbf{x}_j\}_{j=1 \ldots M}}$ be a LiDAR point cloud, where each of the $M$ unordered points ${\mathbf{x}_j \in \mathbb{R}^4}$ consists of the 3D coordinates and the intensity/reflectance value. 
Furthermore, let $F$ be a black-box object detection pipeline and ${F(\mathcal{X})}$ be a set of detections ${\mathcal{D} =  \{\mathbf{d}_k\}}$ to be analyzed.
Our goal is to generate an attribution map ${\Psi_k = \{\psi_{k,j}\}_{j=1 \ldots M}}$ for each $\mathbf{d}_k \in \mathcal{D}$ that reflects the contribution $\psi_{k, j}$ of each point $\mathbf{x}_j \in \mathcal{X}$ to the decision of the model~$F$ \wrt $\mathbf{d}_k$.\\

\noindent\textbf{Attribution Map Generation: } Let $\Omega = \{\omega_j\}_{j = 1 \ldots M}$, $\omega_j \in \{0,1\}$, be a random binary mask. 
Then, we sub-sample $\widetilde{\mathcal{X}} = \{\mathbf{x}_j \in~\mathcal{X} \; | \; \omega_j = 1\}$ according to $\Omega$. 
The detected objects ${F(\widetilde{\mathcal{X}})}$ in this sub-sampled point cloud are given by ${\widetilde{\mathcal{D}}} = \{\widetilde{\mathbf{d}_l}\}$. 
For a single detection ${\mathbf{d}_k} \in \mathcal{D}$ and a single 3D point ${\mathbf{x}_j \in \mathcal{X}}$, 
we then define the contribution ${\psi_{k,j} \in \Psi_k}$ as the expected value of the similarity metric~$S$ over all possible masks ${\Omega}$, where $\mathbf{x}_j$ was part of the input, \ie~${\omega_j = 1}$:
\begin{equation}
\label{eq:method:expectation}
	\psi_{k, j} = \mathbb{E}_{\forall \Omega}\left[\; S\!\left(\mathbf{d}_k, \widetilde{\mathcal{D}}\right) \; | \; \omega_j = 1 \right].
\end{equation}
${S(\mathbf{d}_k, \widetilde{\mathcal{D}})}$ evaluates how well the object described by $\mathbf{d}_k$ was detected despite removing individual points, see Section~\ref{sec:method:metric}. 
For this purpose, we compare $\mathbf{d}_k$  with all detections in ${\widetilde{\mathcal{D}} =  \{\widetilde{\mathbf{d}}_l\}}$, similar to the image-based analysis in~\cite{petsiuk2021CVPR}. 
The intuition behind this idea is that an object can be detected more accurately if important 3D points for the detection are visible. 
The expected value in Eq.~\eqref{eq:method:expectation} can be approximated using Monte Carlo sampling, as shown in~\cite{petsiuk2018BMVC}. 
Therefore, a set of masks ${\{\Omega^i\}_{i=1 \ldots N}}$ is randomly generated from a probability distribution over the domain of possible masks and the corresponding sub-sampled point clouds ${\{\widetilde{\mathcal{X}}^i\}_{i=1 \ldots N}}$ are used as input to the detector. 
The resulting contribution scores are then aggregated to obtain 
\begin{equation}
\label{eq:method:montecarlo}
\psi_{k,j} \approx \frac{1}{\mathbb{E}[\omega_j] \cdot N} \; \sum_{i=1}^{N} S\!\left(\mathbf{d}_k, \widetilde{\mathcal{D}}^i\right) \cdot \omega_{j}^i \;,
\end{equation}
where ${\omega_{j}^i \in \Omega^i}$ indicates the presence of the 3D point $\mathbf{x}_j$ in point cloud ${\widetilde{\mathcal{X}}^i}$. 
The contribution of each point is normalized by the expected number of samples in which the point was visible, \ie~${\mathbb{E}[\omega_j] \cdot N}$. 
We can replace this normalization value by the empirical observation~${\sum_{i=1}^{N} \omega_{j}^i}$.

\subsection{Voxel-based Sampling}
\label{sec:method:voxel-masking}
Sub-sampling influences both, the accuracy of the resulting attribution maps and the number of necessary iterations $N$ to get a good approximation of Eq.~\eqref{eq:method:montecarlo}. 
If we remove too few points, many iterations are necessary to gain information for the whole input. 
Removing large input areas, on the other hand, leads to coarse attribution maps. 
Thus, properly sub-sampling the point clouds is more complex in contrast to images, where the adjacency is well-defined and patches can easily be sampled in a regular~grid. 

Removing the 3D points individually would lead to a huge space of possible masks $\Omega$ and $N$ would be impracticably high. 
In addition, the ability to evaluate the impact of hiding entire areas, such as the side of a vehicle, would be limited: 
The probability that all points of a specific area are removed while other areas remain untouched within a single mask would be very low.

Therefore, we propose to subdivide the 3D space along the $x$, $y$ and $z$ axis into uniformly distributed voxels of equal edge length. 
We then group the points according to the voxel in which they are located. 
In each iteration, all points of the $v^\text{th}$ voxel are kept jointly with probability $P_v$, by setting the corresponding mask entries to ${\omega_j=1}$, or removed jointly, {$\omega_j=0$}, with the counter probability ${1-P_v}$. 
We argue that the pattern of several neighboring points being removed simultaneously is also very common in real LiDAR data, \eg when an object is partially occluded by another object. 
To avoid always grouping the same points into the same voxel in each iteration, the voxel grid is randomly rotated and translated in all directions.

This sub-sampling approach has two advantages: First, by grouping the points into voxels we drastically reduce the number of necessary iterations $N$. Second, by randomly changing the voxel grid in each iteration it is still possible to obtain point-level insights.

\subsection{Density-aware Sampling Strategy}
\label{sec:method:density-aware}
\begin{figure}
	\centering
	\includegraphics[width=0.95\linewidth]{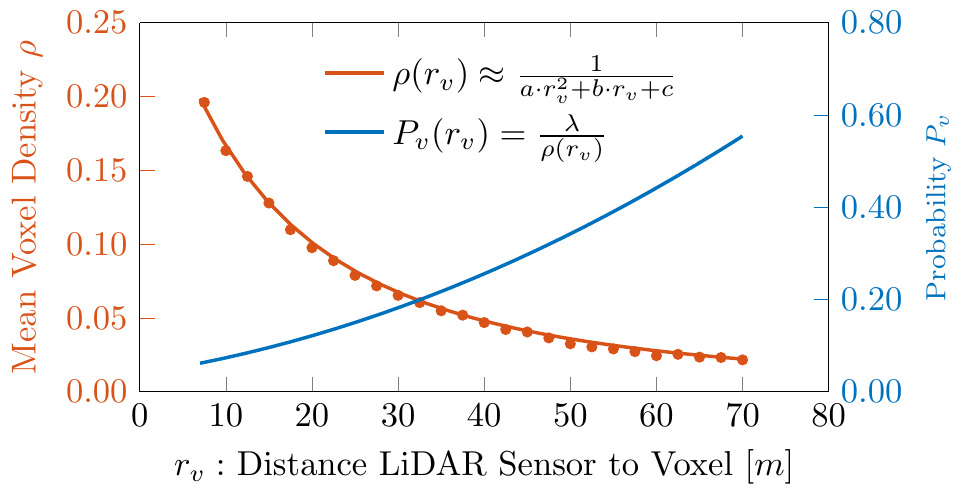}
	\caption{Visualization of the mean voxel density, $\rho \in [0,1]$, for the KITTI dataset~\cite{geiger2012CVPR} (voxel size~${20 \; \text{cm}}$), defined as the mean percentage of non-empty voxels within ${1 \; \text{m}}$,  depending on the distance $r_v$. 
		We approximate this density by a second order polynomial $\rho (r_v)$ and use its inverse, scaled by the hyperparameter $\lambda$, to determine the sampling probability $P_v$ for each voxel.}
	\label{fig:density}
\end{figure}

As shown in Eq.~\eqref{eq:method:montecarlo}, each mask is weighted by how accurately the object of interest was detected in the sub-sampled point cloud. 
Therefore, the quality of the final attribution map depends on how well the detector was challenged. 
If the probability $P_v$ to keep the relevant voxels and corresponding points is too low and the object of interest is never detected, no information can be obtained. 
The same is true if $P_v$ is too high and the object is always detected accurately. 
We must also take into account that the density of the points is not uniformly distributed, in contrast to the pixels in an image. 
The point density depends strongly on the distance to the LiDAR sensor. This means that for close objects, more voxels have to be removed than for distant objects to challenge the detector appropriately. 
From Figure~\ref{fig:density}, we see that a car placed at a distance of ${10 \; \text{m}}$ consists, on average, of about 4 times as many voxels as a car placed ${50 \; \text{m}}$ away.
Thus, the contribution of a single voxel to the model's decision increases with the distance from the LiDAR sensor. 
More formally, the distance between two neighboring laser rays that hit a surface orthogonal to the LiDAR sensor is given by ${r_{\text{obj}} \cdot \tan(\alpha)}$, where $r_{\text{obj}}$ is the distance to the object and $\alpha$ the angular resolution of the LiDAR sensor. 
Thus, assuming a square area of a certain size, the number of data points hitting that area decreases quadratically with the distance from the LiDAR sensor. 

Therefore, we approximate the mean voxel density once for a dataset with a second order polynomial~${\rho(r_v) = (a \cdot r_v^2 + b \cdot r_v + c)^{-1}}$. 
When generating the attribution maps, we use the scaled inverse of this mean density to determine $P_v$ for each voxel,
\begin{equation}
	P_{v}(r_v) = \lambda \frac{1}{\rho(r_v)},
\end{equation}
where $r_v$ is the distance from the LiDAR sensor to the voxel center and $\lambda$ is a hyperparameter. 
The probability $P_v$ for each voxel thus depends on the mean voxel density~$\rho$, scaled by $\lambda$ to adjust the desired density in the sub-sampled point cloud. 
The relationship between the mean voxel density~$\rho$ and  $P_v$ is also shown in Figure~\ref{fig:density}. 
This sub-sampling strategy enables us to challenge the detector appropiately, regardless of the distance from the LiDAR sensor.

\subsection{Similarity Metric}
\label{sec:method:metric}
The similarity metric~$S$ measures how well the (presumed) object~$\mathbf{d}_k$ was still detected after sub-sampling the input. 
For this purpose, $\mathbf{d}_k$ is compared with all detections ${\widetilde{\mathcal{D}} =  \{\widetilde{\mathbf{d}}_l\}}$ from the sub-sampled point cloud. 
Inspired by the image-based approach~\cite{petsiuk2021CVPR}, we define $S$ as maximum pairwise similarity between $\mathbf{d}_k$ and the detections~$\widetilde{\mathbf{d}}_l \in \widetilde{\mathcal{D}}$, 
\begin{equation}
	S(\mathbf{d}_k, \widetilde{\mathcal{D}}) = \max_{\mathbf{\widetilde{d}}_l \in \widetilde{\mathcal{D}}} \; s(\mathbf{d}_k, \widetilde{\mathbf{d}}_l),
\end{equation}
and adapt it to the requirements of 3D object detection on LiDAR data. 
Each detection $\mathbf{d}_k$ is defined by a class label $y_k$, a confidence score $c_k$ and a 3D bounding box $\mathbf{b}_k$. 
To compute the pairwise similarity between two detections,
\begin{equation}
	s(\mathbf{d}_k, \mathbf{\widetilde{d}}_l) = \prod_{A \in \mathbb{A}} \; s_{A}(\mathbf{d}_k, \mathbf{\widetilde{d}}_l),
\end{equation}
we define a set of sub-metrics $\mathbb{A}$ to determine the individual differences in the label, the confidence score, and the consensus of the bounding boxes. 
The first two sub-metrics $s_{\text{class}}$ and $s_{\text{overlap}}$ ensure that only detections of the same class and overlapping detections, IoU $>$ 0, can lead to a non-zero similarity:
\begin{align}
s_{\text{class}}(\mathbf{d}_k, \mathbf{\widetilde{d}}_l) &=
\begin{cases}
1        & \text{if} \quad y_k = \widetilde{y}_l \\
0        & \text{otherwise},
\end{cases}
\\
s_{\text{overlap}}(\mathbf{d}_k, \mathbf{\widetilde{d}}_l) &=
\begin{cases}
1  & \text{if} \quad \text{IoU}(\mathbf{b}_k, \mathbf{\widetilde{b}}_l) > 0 \\
0  & \text{otherwise}.
\end{cases}
\end{align}
To ensure that a potential increase in the confidence score compared to the original detection, \ie~$\widetilde{c}_l > c_k$, does not lead to a penalization, we directly use~$\widetilde{c}_l$ for
\begin{equation}
	s_{\text{conf}}(\mathbf{d}_k, \mathbf{\widetilde{d}}_l) = \widetilde{c}_l.
\end{equation}
To determine the correspondence of the pairwise 3D bounding boxes as precisely as possible, we evaluate the rotation, translation and scaling separately, inspired by the true positive metrics of nuScenes~\cite{caesar2020CVPR}. 
Each bounding box $\mathbf{d}$ is given by its center coordinates $\mathbf{C}$, its box dimensions and its yaw angle $\theta$. 
The translation similarity is computed as the Euclidean distance between the bounding box centers,
\begin{equation}
	s_{\text{translation}}(\mathbf{d}_k, \mathbf{\widetilde{d}}_l) = \max \left(1 - \|\mathbf{C}_k - \widetilde{\mathbf{C}}_l\|,\; 0\right).
\end{equation}
For the scale similarity, we align the bounding boxes \wrt their center and yaw angle and then compute their IoU,
\begin{equation}
	s_{\text{scale}}(\mathbf{d}_k, \mathbf{\widetilde{d}}_l) = \text{IoU} \; (\mathbf{b}_k^{\text{aligned}}, \mathbf{\widetilde{b}}_l^{\text{aligned}}).
\end{equation}
Finally, the orientation similarity is determined by the smallest yaw angle difference between $\theta_k$ and $\widetilde{\theta}_l$ in radians,
\begin{equation}
s_{\text{orientation}}(\mathbf{d}_k, \mathbf{\widetilde{d}}_l) = \max (1 - |\theta_k - \widetilde{\theta}_l|,\; 0).
\end{equation}
The advantage of these separate scores is that the localization of the object can be evaluated more accurately than by using only the IoU. 
Additionally, this allows the creation of attribution maps for individual properties.

\section{Experiments}
\label{sec:experiments}

We demonstrate OccAM's insights into the behavior of a PointPillars~\cite{lang2019CVPR} model in detail and compare the behaviors across several recent detectors, including SECOND~\cite{yan2018Sensors} and PV-RCNN~\cite{shi2020CVPR:2}.

\subsection{Implementation Details}
Our experimental setup\footnote{\url{https://github.com/dschinagl/occam}} is based on the open-source toolbox OpenPCDet~\cite{pcdet2020SW}. 
Unless otherwise stated, all detectors are trained on the KITTI dataset~\cite{geiger2012CVPR} with their default configurations and using the standard training policies. 
We use the standard split~\cite{chen2015NIPS} into training and validation samples. 
All shown attribution maps are exclusively from the 3769 samples of the validation set, which we also use to estimate the voxel density $\rho$. We use a voxel size of ${20\;\text{cm}}$ (cube-shaped) in all our experiments.

We determine the values of the hyperparameters $N$ and $\lambda$ empirically. 
We found that $N=3000$ is a good compromise between the runtime and the quality of the attribution maps (details in supplemental material). 
The parameter $\lambda$ was chosen such that at a distance of ${r_v = 25\;\text{m}}$ the sampling probability $P_v$ is equal to $0.15$, unless explicitly stated. 
This configuration gives stable results regardless of the detector's architecture.

Since all $N$ probing steps are independent of each other, the entire process is fully parallelizable. 
Furthermore, the perturbed point clouds can be efficiently generated on the CPU. 
The runtime is thus dominated by the detector runtime on the GPU. 
For PointPillars, the analysis of a KITTI point cloud takes on average ${50\,\text{s}}$ (for ${N=3000}$) on a desktop PC using a single NVIDIA\textsuperscript{\textregistered} GeForce\textsuperscript{\textregistered} RTX 3090 GPU.

\subsection{Qualitative Results}
\begin{figure}
	\centering
	\includegraphics[width=1\linewidth]{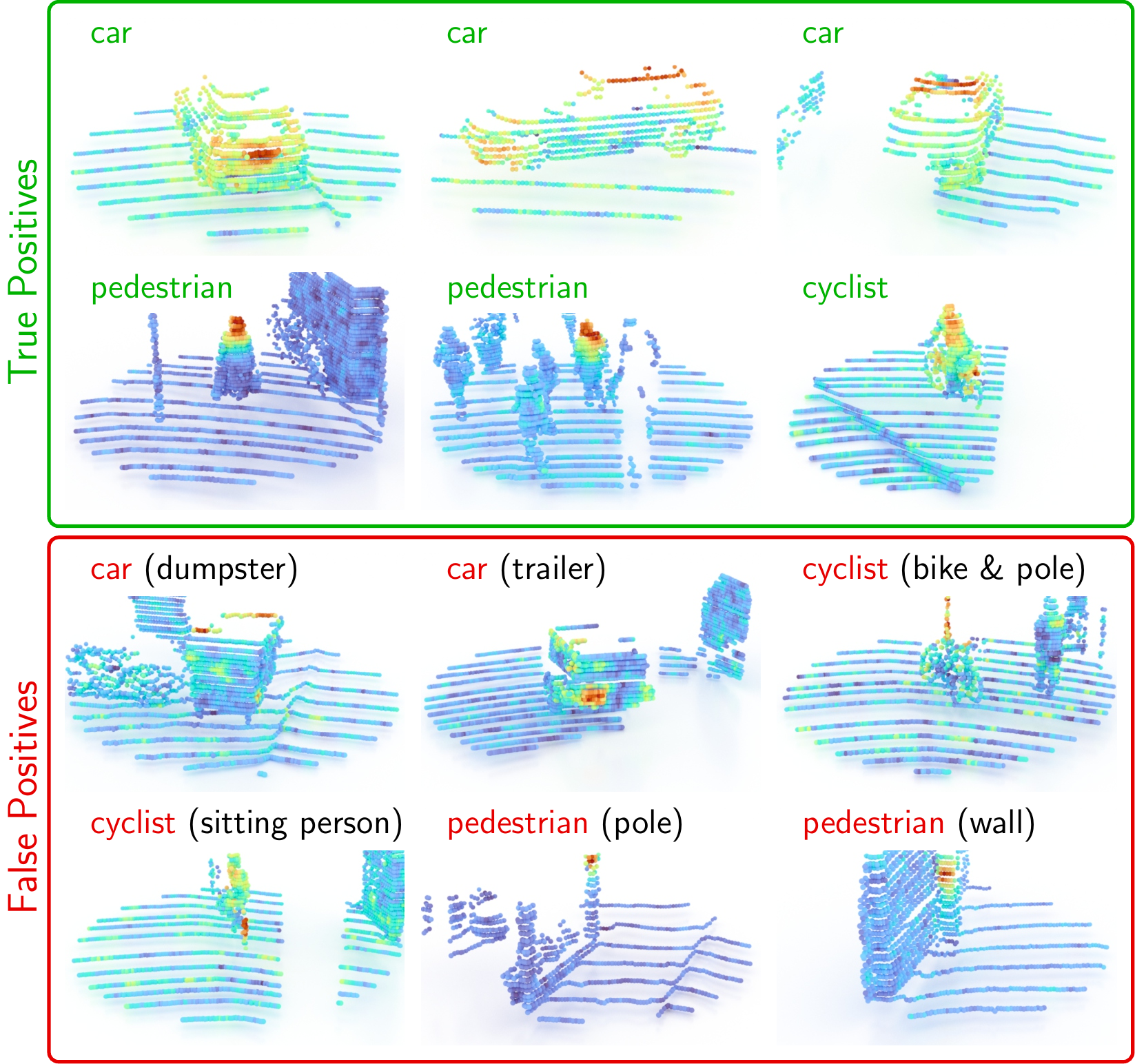}
	\caption{Attribution map examples for PointPillars detections on KITTI. 
		Warmer colors (turbo colormap) denote higher contribution of a point to this detection. Crops are for visualization only.}
	\label{fig:am_samples}
\end{figure}
We demonstrate that our attribution maps are interpretable and informative. 
Figure~\ref{fig:am_samples} shows examples for both, true and false positive detections. 
Considering the \emph{car} class we have found that the license plates, due to their high reflectance, represent a very discriminative feature. 
Furthermore, the roof railing, the A-pillar and the B-pillar are very distinct features of this class. 
We argue that these areas have a high identification value due to their shape and the transparency of the windows. 
Especially, if a car is only seen from the side or the license plate is occluded by other objects, these areas are very important for detection. 
For \emph{pedestrians}, as expected, the head-shoulder silhouette proved to be a highly discriminatory feature and was consistently shown as the most important area. 
This area is also very relevant for the detection of \emph{cyclists} in addition to the reflectors on the front and rear of the bike. 
The high interpretability is also evident for false detections. 
For example, in the case of the mis-detection of the dumpster as a vehicle, the top-rear area was obviously incorrectly interpreted as roof railing. 
In the case of the false detection of the trailer, the detector was obviously misled by the license plate. 
Note also how well our attribution maps allow us to analyze single objects even in crowded scenes, \eg~see the pedestrian in the group. 
These examples show that our attribution maps can provide useful insights into the model behavior and are also well suited to analyze false detections.

\begin{figure*}
	\centering
	\subfloat[\label{fig:average_am} Average attribution maps (turbo-colored) for PointPillars~\cite{lang2019CVPR} trained and evaluated on KITTI~\cite{geiger2012CVPR}. 
	We also show the average LiDAR reflectivity~/~intensity values (copper-colored). 
	From left to right: cars, pedestrians and cyclists.]{\includegraphics[width=16cm]{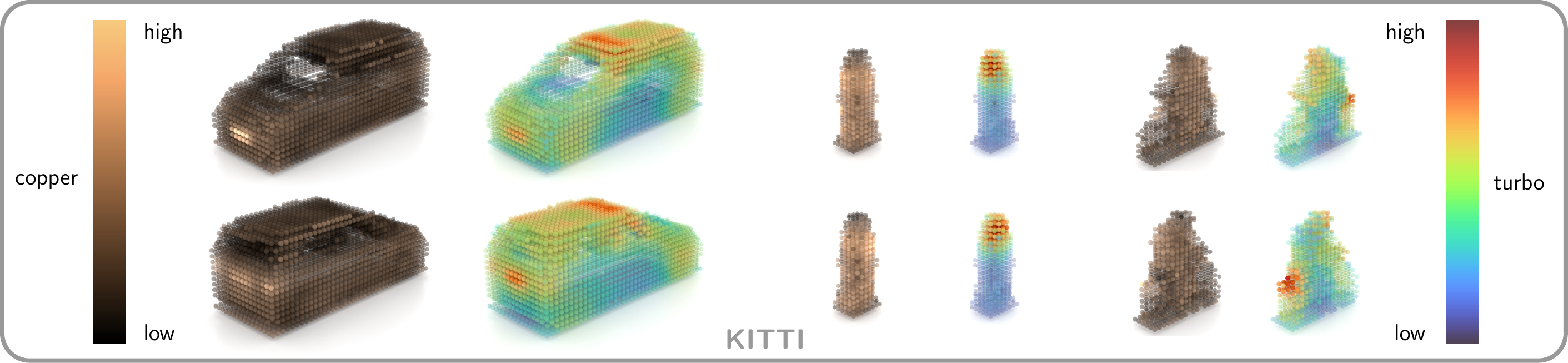}}\\ \vspace{5mm}
	\subfloat[\label{fig:comparison} Comparison of PointPillars~\cite{lang2019CVPR} (trained with and without reflectivity values), SECOND~\cite{yan2018Sensors} and PV-RCNN~\cite{shi2020CVPR:2} trained and evluated on KITTI~\cite{geiger2012CVPR}. 
	Note how the average attribution maps reflect the granularity of the underlying detector architecture.]{\includegraphics[height=4.525cm]{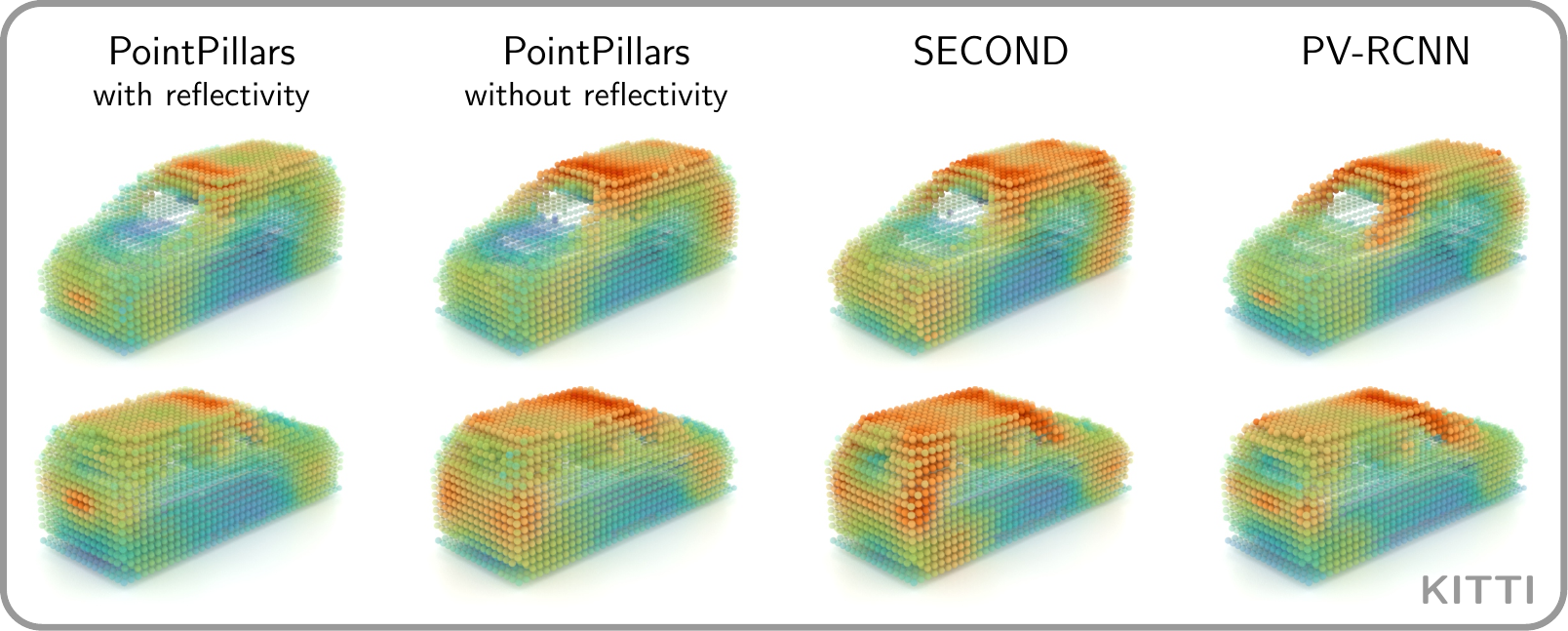}} \hspace{2mm} 
	\subfloat[\label{fig:waymo} SECOND~\cite{yan2018Sensors} trained and evaluated on the Waymo Open Dataset (WOD)~\cite{sun2020CVPR}. 
	For visualization purposes, we clustered the vehicles by size.]{\includegraphics[height=4.525cm]{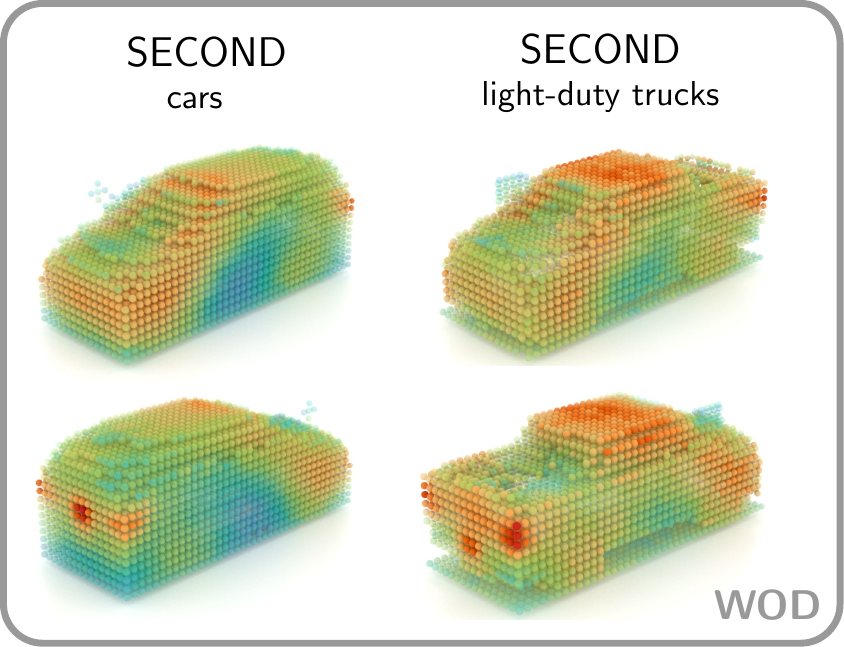}}
	\caption{Average attribution maps (turbo-colored) on \subref{fig:average_am} \subref{fig:comparison} KITTI~\cite{geiger2012CVPR} and, for comparison, on \subref{fig:waymo}~Waymo Open Dataset (WOD)~\cite{sun2020CVPR}. 
		We adjust the opacity of the voxels according to their attribution score. Best viewed on screen.}
	\label{fig:short}
\end{figure*}

\subsection{Average Attribution Maps}
To verify these findings from the individual attribution maps, we also average the maps of the specific classes. 
To do so, we first scale all boxes and the associated points to a uniform size and then align them \wrt their center and yaw angle. 
We then voxelize the resulting point cloud and average the attribution values of the individual points within a voxel. 
The resulting maps are shown in Figure~\ref{fig:average_am}. 
Our previous findings from the individual attribution maps, such as the importance of license plates or the head-shoulder silhouette, are also clearly visible in these averaged maps.

\subsection{Metrics}
Evaluating the explainability of attribution maps is still an open question, even for image-based analysis methods.
We adopt the \emph{point dropping} approach~\cite{zheng2019ICCV} for point cloud classification models.
The basic idea is to iteratively remove the most important points and re-evaluate the model \wrt IoU and confidence.
If the attribution maps are expressive, then the model's performance should degrade faster than by just removing points randomly.
On the contrary, if the least important points are removed first, then the performance should degrade slower than during random removal.

For 3D object detection, however, this idea unfairly penalizes the random removal baseline.
Thus, we apply point dropping only to points within actual detections.
For this we use the original black-box detector output as pseudo ground truth.
Iteratively, we then remove the most/least important points, re-run the detector and compare its output to the pseudo ground truth.
Since the random baseline also removes points only within detections, this is a fair comparison: these points are obviously more important than unrelated points outside, far-away from any detection.

\begin{figure}
	\centering
	\includegraphics[width=1\linewidth]{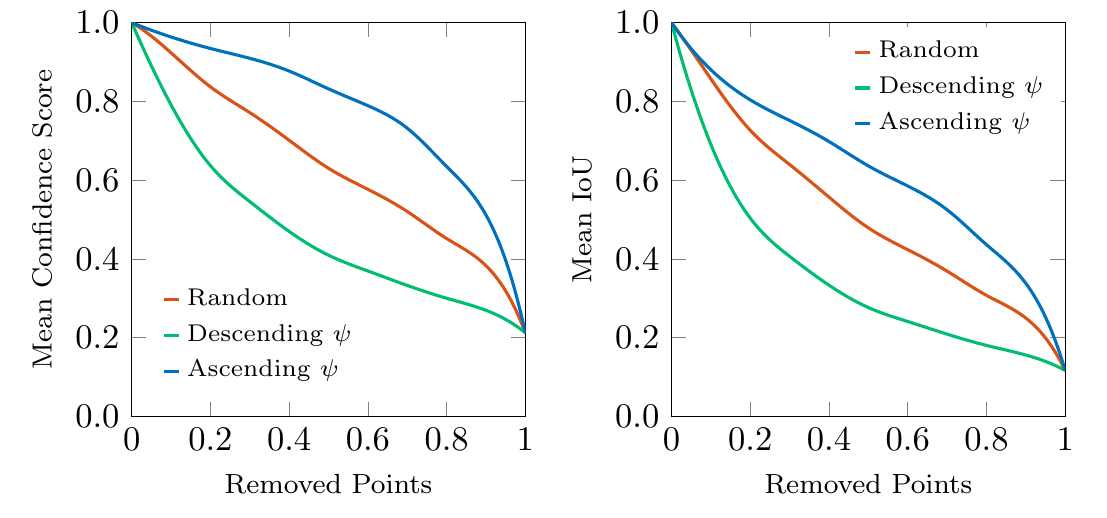}
	\caption{ Point dropping evaluation. 
		Points within each bounding box are removed in descending or ascending order of importance, and the resulting detections are compared with the original detections using confidence score and IoU.}
	\label{fig:eval}
\end{figure}

\begin{figure}
	\centering
	\includegraphics[width=1\linewidth]{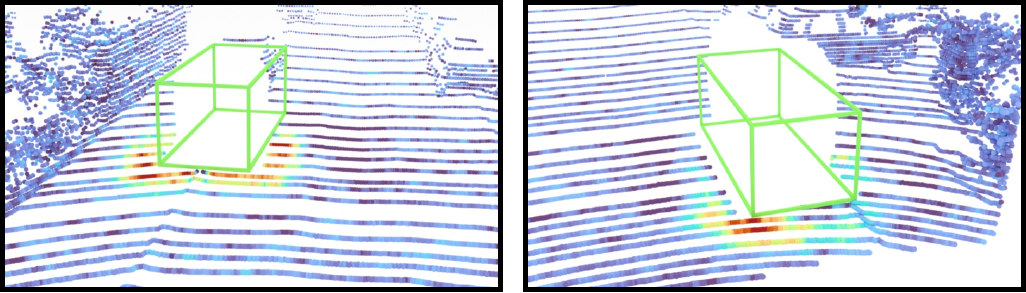}
	\caption{Detections despite empty bounding boxes: The attribution maps show that distinctive cutouts on the ground are responsible for these PointPillars detections. 
		Since detections with no points are obviously more sensitive to perturbation, we set $\lambda$ s.t. $P_v = 0.3$ at $r_v=25\, \text{m}$ for these visualizations.}
	\label{fig:emptyboxes}
\end{figure}

As shown in Figure~\ref{fig:eval}, removing the most/least important points first leads to a significantly faster/slower model degradation in contrast to random removal.
Thus, our attribution maps are expressive.
Note that both IoU and the confidence score do not reach zero, even if all points within the bounding boxes are removed.
This is caused by the interesting finding that surrounding features also notably contribute to the detector's decision, \eg~see Figure~\ref{fig:emptyboxes}.

\subsection{Importance of Density-aware Sampling}

\begin{figure}
	\centering
	\includegraphics[width=1\linewidth]{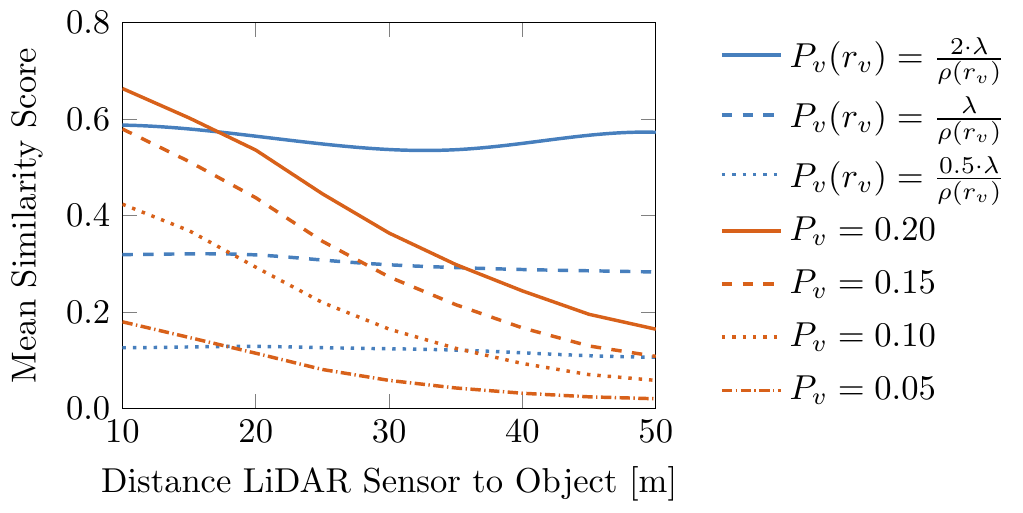}
	\caption{Different sampling schemes for a PointPillar model. We plot the mean similarity score over the distance between the object of interest and the LiDAR sensor. 
		With a fixed $P_v$, attribution maps would only be expressive in a narrow distance range.}
	\label{fig:lambda_eval}
\end{figure}

To demonstrate the importance of our density-aware sampling strategy, we compare our point density-derived distance-dependent choice of $P_v(r_v)$ with a fixed choice for~$P_v$. 
Figure~\ref{fig:lambda_eval} shows the mean similarity score during the attribution map computation as a function of the distance between the object of interest and the LiDAR sensor. 
With a fixed choice of $P_v$ the mean similarity score strongly depends on the distance to the object. 
Intuitively, close objects can be detected more accurately than distant objects. 
However, if the objects are recognized too precisely or only very rarely during this probing phase, these samples do not lead to any information gain. 
This means that for fixed $P_v$ there is only a small distance range in which meaningful attribution maps would be created.

In contrast, our density-aware sampling strategy leads to an almost constant similarity score independent of the distance to the object. 
This allows the computation of attribution maps for all objects regardless of their distance to the sensor with only a single probing process.

\subsection{Comparison of Detector Architectures}
The attribution maps generated by OccAM are well suited to compare the behavior of different object detectors. 
In Figure~\ref{fig:comparison}, we show the average car attribution maps for PointPillars~\cite{lang2019CVPR}, SECOND~\cite{yan2018Sensors} and PV-RCNN~\cite{shi2020CVPR:2}. 
For pedestrians and cyclists we found out that the head-shoulder silhouette is the most important feature, regardless of the analyzed architecture. 
These additional maps and further detectors are included in the supplemental material.

For all detectors, the points on the roof are an important feature to distinguish a vehicle from other classes while the door areas are of minor relevance. 
Another distinct characteristic of cars are the well-reflecting license plates and lights. 
Here, however, the detectors differ: while PointPillars focuses  on the license plates, these are less important for SECOND. 
We also trained PointPillars without reflectivity values, which then focuses more on the edges of the object, similar to SECOND. 
For PV-RCNN, the vehicle's license plates and lights appear to be of balanced importance.

The influence of the underlying architecture and the point cloud processing method can best be seen at the A-pillars. 
Due to the transparent windows of the car, the A-pillar represents a very distinct feature. 
While this feature is of little importance for PointPillars, due to its coarse (pillar) processing, it is notably more important for the voxel-based SECOND model. 
PV-RCNN leverages these fine structures best, since it processes point and voxel information jointly.

To further demonstrate OccAM's general applicability, we provide the average attribution maps of SECOND on the Waymo Open Dataset (WOD)~\cite{sun2020CVPR} in Figure~\ref{fig:waymo}.
Since, in contrast to KITTI, WOD has only a single \emph{vehicle} class (which also includes trucks, \emph{etc.}), we cluster the individual maps by vehicle size before averaging.
These maps show both, similarities (focus on roof and A-pillars) and differences (varying importance license plates, lights and fender regions) between KITTI~(Figure~\ref{fig:comparison}) and WOD~(Figure~\ref{fig:waymo}).

\section{Conclusion}
\label{sec:conclusion}
We proposed OccAM, the first method to generate attribution maps for black-box 3D object detectors on LiDAR data. 
Our attribution maps are well interpretable, informative and provide useful insights into the behavior of the detectors. 
While some behavior was already anticipated, \eg head-shoulder silhouette, we also found unexpected insights, \eg the importance of the A-pillars and the local surroundings. 
With OccAM, we offer the community a way to analyze the behavior of any 3D object detector on any LiDAR dataset.

{
\small
\noindent\textbf{Acknowledgements}
This work was partially funded by the Austrian FFG project iLIDS4SAM~(878713) and by the Christian Doppler Laboratory for Embedded Machine Learning. 
Peter M. Roth received support from the German Federal Ministry of Education and Research (BMBF) within the international future AI lab AI4EO (01DD20001).
}

{\small
\bibliographystyle{ieee_fullname}
\bibliography{main.bib}
}
\newpage
\appendix
\section{Supplemental Material}

This supplementary discusses additional insights, quantitative results and potential limitations of our approach.

\subsection{Detector Comparison}

\begin{figure*}
	\centering
	\includegraphics[width=1\linewidth]{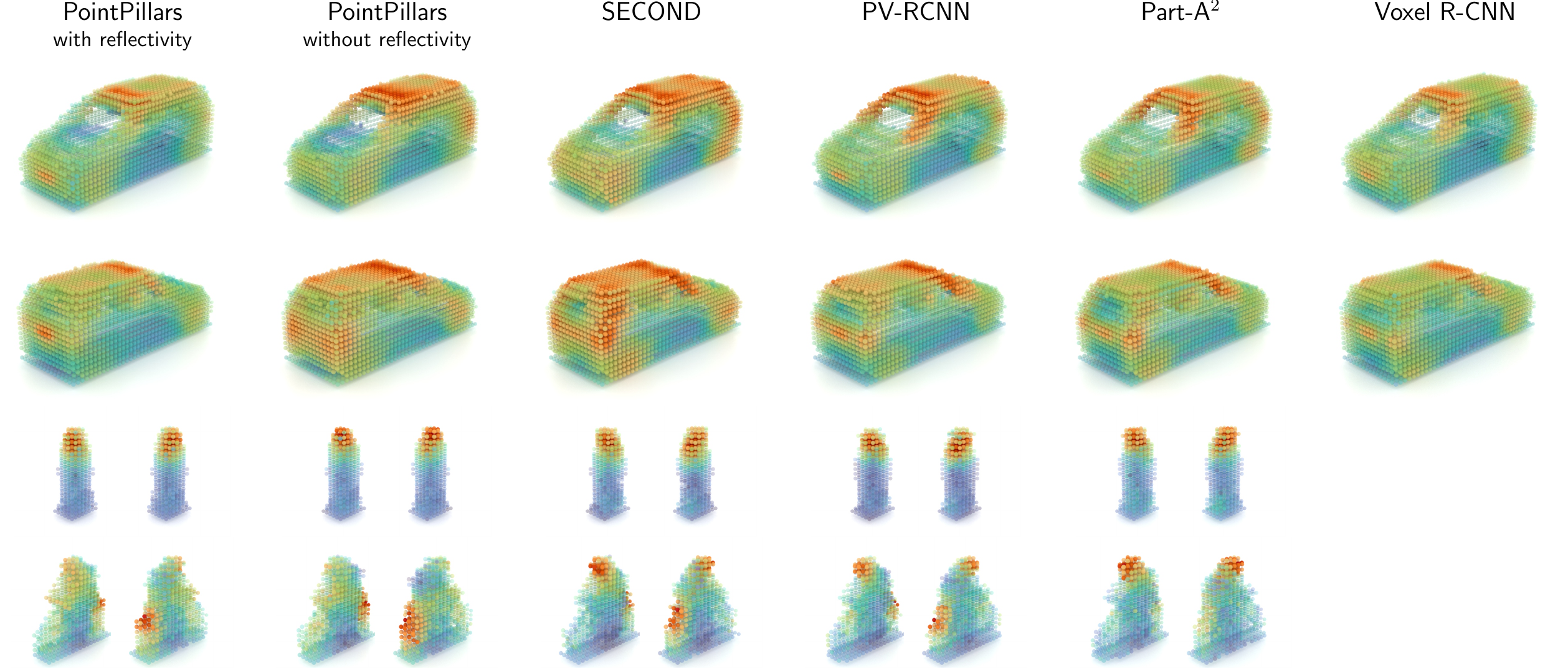}
	\caption{Comparison of average attribution maps for PointPillars~\cite{lang2019CVPR} (trained w/o reflectivity), SECOND~\cite{yan2018Sensors}, PV-RCNN~\cite{shi2020CVPR:2}, Part-A²~\cite{shi2021PAMI} and Voxel R-CNN~\cite{denk2021AAAI} trained \& evaluated on KITTI~\cite{geiger2012CVPR}. 
		Note: Voxel R-CNN (from  OpenPCDet~\cite{pcdet2020SW} model zoo) is only available for cars.}
	\label{fig:comparison_suppl}
\end{figure*}

Due to the page limit, we omitted the average attribution maps for pedestrians and cyclists, as well as Part-A$^2$~\cite{shi2021PAMI} (in favor of the also hybrid approach PV-RCNN~\cite{shi2020CVPR:2}) and Voxel R-CNN~\cite{denk2021AAAI} (in favor of the also voxel-based SECOND~\cite{yan2018Sensors}) from the main manuscript (Section 4.6). 
Figure~\ref{fig:comparison_suppl} shows these maps for all classes and all detectors.

Considering pedestrians, we see that independent of the model's architecture the head-shoulder silhouette is the most important feature for detection. 
In the case of cyclists, the voxel-based and hybrid methods show a stronger focus on the head than the pillar-based method. 
Furthermore, we can see that the two hybrid methods PV-RCNN~\cite{shi2020CVPR:2} and Part-A$^2$~\cite{shi2021PAMI} lead to very similar average attribution maps.

\subsection{Number of Iterations}
\begin{figure*}
	\centering
	\subfloat[\label{fig:N_eval_ind} Detection-specific attribution maps from top-to-bottom: pedestrian, car, cyclist, car.]{\includegraphics[width=0.965\linewidth]{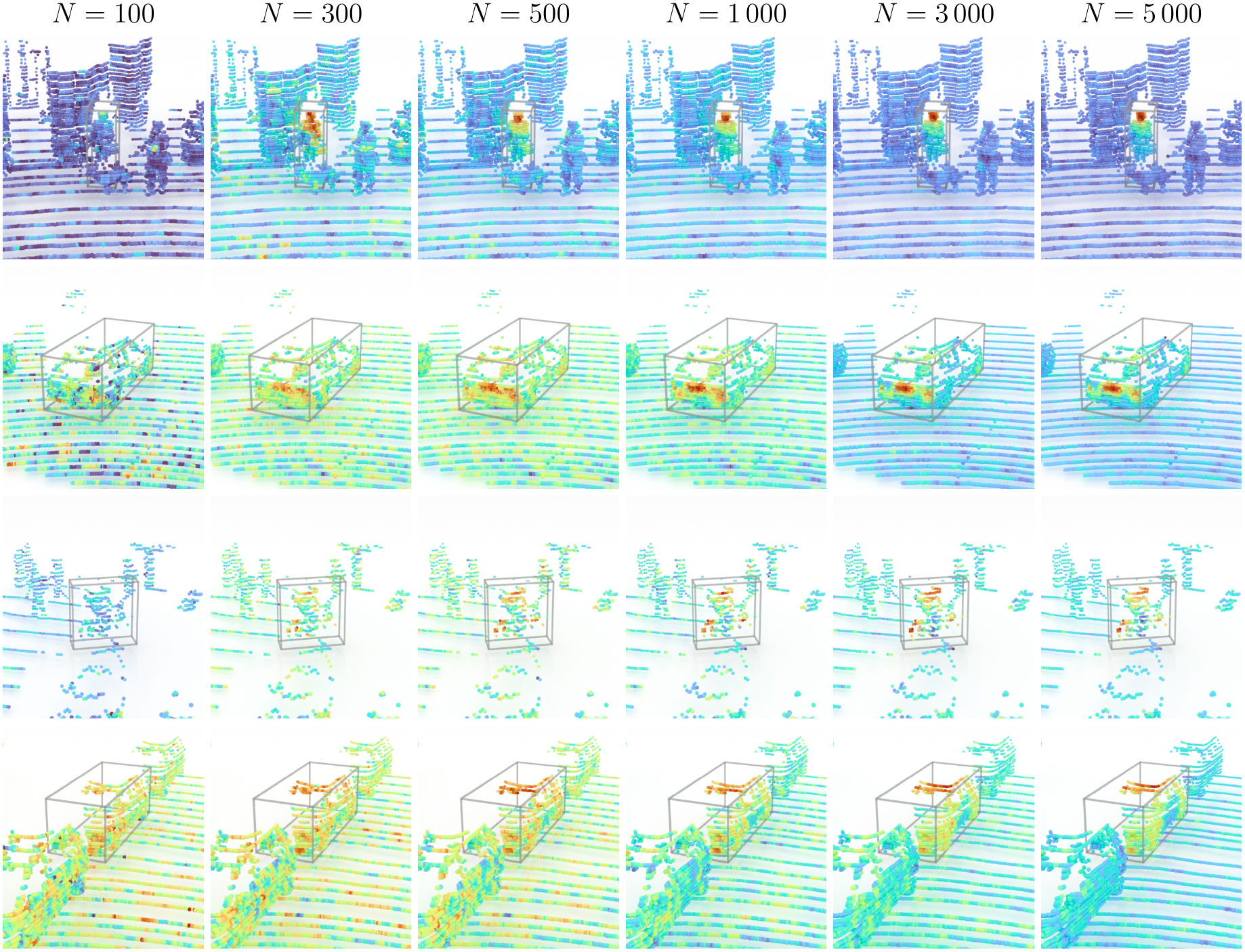}}\\ \vspace{5mm}
	\subfloat[\label{fig:average_am_evalN} Average attribution maps.]{\includegraphics[width=0.965\linewidth]{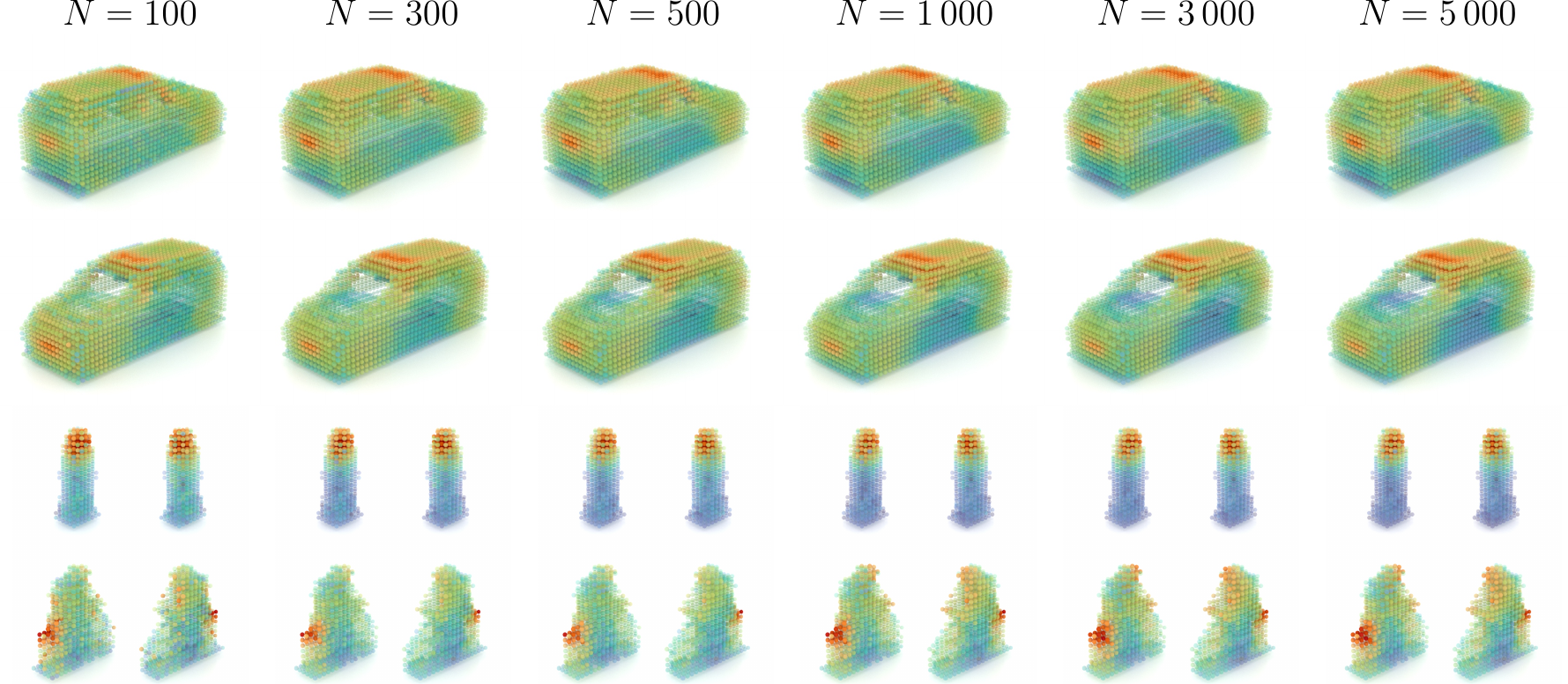}}
	\caption{Influence of the number of iterations $N$ on \subref{fig:N_eval_ind} individual (\ie per-detection) and \subref{fig:average_am_evalN} average attribution maps for PointPillars~\cite{lang2019CVPR} trained and evaluated on KITTI~\cite{geiger2012CVPR}. Best viewed on screen.}
	\label{fig:N_eval}
\end{figure*}

Figure~\ref{fig:N_eval_ind} shows how the number of iterations $N$ influences the quality of individual (\ie per-detection) attribution maps.
While the attribution maps are very noisy for $N=100$, the importance of individual regions can already be seen at $N=300$ (which takes on average only $6$~seconds for PointPillars~\cite{lang2019CVPR}).
Further iterations allow us to refine the details of the attribution maps notably.
We found that the quality (regardless of the detector) saturates around $N=3000$ and thus, use this setting for all experiments (as stated in the main manuscript, Section 4.1).

Considering the average attribution maps, shown in Figure~\ref{fig:average_am_evalN}, 
we notice that the most important regions are already recognizable very early on.
This is due to the fact that averaging the individual attribution maps is analogous to increasing $N$ as above.
Nevertheless, additional iterations further improve the precision of these maps.

\subsection{Convergence}
\begin{figure}[!t]
	\centering
	\includegraphics[width=0.9\linewidth]{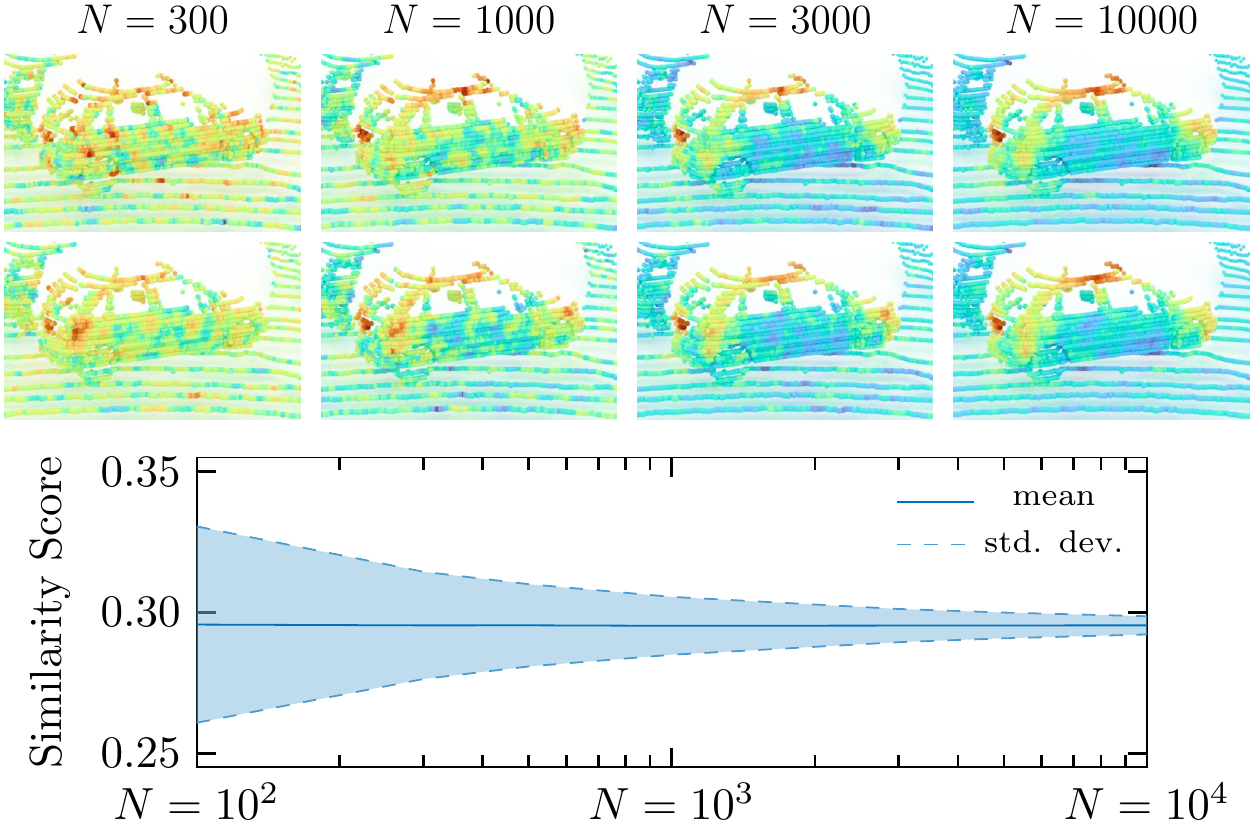}
	\caption{
		Convergence of the similarity score per point (for 500 detections over 20 runs with different random seeds). 
		The visuals on top show the attribution map progression for a \emph{car} over 2 runs.
	}
	\label{fig:variancesamples}
\end{figure}

To demonstrate that our attribution maps converge towards the same result, we show in  \cref{fig:variancesamples} 
(bottom) the mean similarity score and std. dev. for 500 detections over 20 runs with different random seeds (computed for all points within detections). 
In addition we provide an attribution map progression for a car over 2 differently initialized runs.

\subsection{Individual Similarity Sub-Metrics}

\begin{figure*}
	\centering
	\includegraphics[width=0.83333\linewidth]{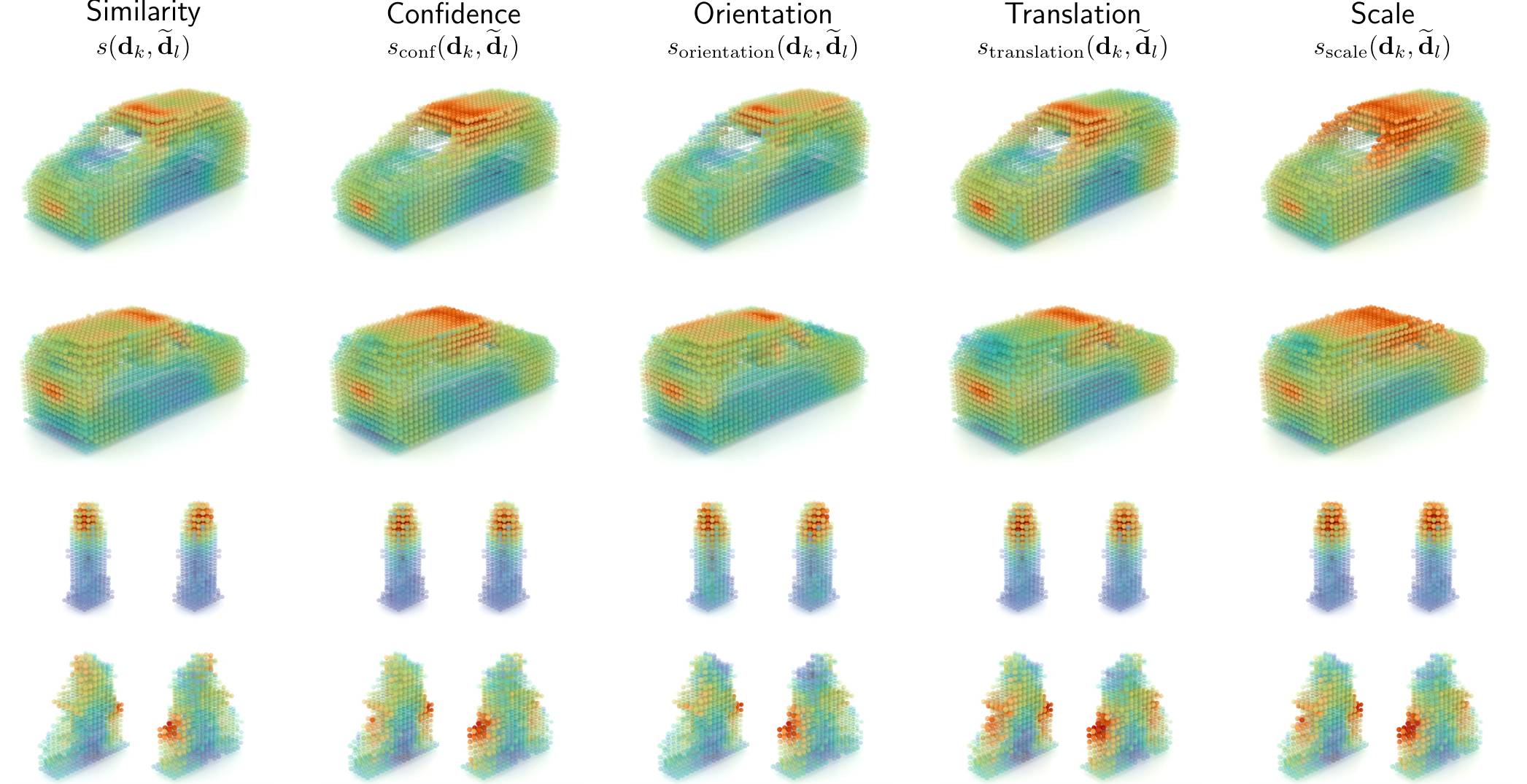}
	\caption{Average attribution maps \wrt the individual sub-metrics for a PointPillars~\cite{lang2019CVPR} model trained and evaluated on KITTI~\cite{geiger2012CVPR}.}
	\label{fig:average_am_score_separate}
\end{figure*}

Our similarity metric allows the generation of attribution maps for individual sub-metrics (\emph{cf.} main manuscript Section~3.3). 
Figure~\ref{fig:average_am_score_separate} shows the average attribution maps \wrt orientation, translation, scale and confidence score compared to the overall similarity score. 
These show that (especially for cars) different structures are of varying importance for the detector's decision. 
For example, estimating the orientation focuses clearly on the roof, whereas for proper scaling the A-pillars are of higher importance.

\subsection{Pointing Game}

Another metric to assess saliency maps of image-based classification models is the pointing game~\cite{zhang2018IJCV}. 
For each saliency map the position of the max. value is determined. 
If this pixel is within the segmentation mask of the object, the example is considered a hit. 
The \emph{pointing game score} is the number of hits divided by the number of evaluated image samples. 
We check for each correctly detected object whether the maximum attribution map value is within the ground truth 3D bounding box. 
For a PointPillars~\cite{lang2019CVPR} model trained and evaluated on KITTI~\cite{geiger2012CVPR}, we achieve a pointing game score of $0.9004$. 
Note, however, that in KITTI the side mirrors of a car are not part of the annotated bounding box. 
If we thus increase the dimensions of the boxes by only 10~\%, the score increases to $0.9724$.

\subsection{Limitations}
The first potential limitation is the runtime, which is primarily determined by the inference speed of the detector and by the number of iterations $N$. 
Note, however that creating very precise attribution maps is only needed for very few select examples, \eg to analyze mis-detections. 
Creating the average attribution maps, on the other hand, can be done efficiently with a significantly smaller number of iterations. 
A second potential limitation can occur if an object is close to the LiDAR sensor, but still consists of only a few points. 
These edge cases can be caused by severe occlusions by other objects. 
In such cases the detections are more sensitive to the sub-sampling and the hyperparameter $\lambda$ should be adapted individually, as we showed for the empty bounding boxes in Section~4.4 of the main manuscript.

\end{document}